\documentclass[conference,a4paper,9pt]{IEEEtran}
\usepackage{cite}
\usepackage{url}
\usepackage{amsmath}
\usepackage{graphicx}
\usepackage{pgfplots,tikz}
\pgfplotsset{compat=newest}
\usetikzlibrary{plotmarks}
\usetikzlibrary{arrows.meta}
\usepgfplotslibrary{patchplots}
\usepackage{grffile}
\usepackage[hidelinks]{hyperref}

\newcommand\copyrighttext{%
	\footnotesize \textcopyright 2021 IEEE. Personal use of this material is permitted.
	Permission from IEEE must be obtained for all other uses, in any current or future
	media, including reprinting/republishing this material for advertising or promotional
	purposes, creating new collective works, for resale or redistribution to servers or
	lists, or reuse of any copyrighted component of this work in other works.
	DOI: \href{https://doi.org/10.1109/ISSCS52333.2021.9497438}{10.1109/ISSCS52333.2021.9497438}}
\newcommand\copyrightnotice{%
	\begin{tikzpicture}[remember picture,overlay]
		\node[anchor=south,yshift=10pt] at (current page.south) {\fbox{\parbox{\dimexpr\textwidth-\fboxsep-\fboxrule\relax}{\copyrighttext}}};
	\end{tikzpicture}%
}

\begin{document}

\title{3D Rendering Framework for Data Augmentation in Optical Character Recognition}


\author{\authorblockN{Andreas Spruck\authorrefmark{1},
Maximiliane Hawesch\authorrefmark{1},
Anatol Maier\authorrefmark{2},
Christian Riess\authorrefmark{2},
Jürgen Seiler\authorrefmark{1}, and
André Kaup\authorrefmark{1}}
\authorblockA{\authorrefmark{1}Chair for Multimedia Communications and Signal Processing\\
University Erlangen - Nürnberg,\\
Erlangen, Germany\\}
\authorblockA{\authorrefmark{2}Chair of Computer Science 1 (IT Security Infrastructures)\\
University Erlangen - Nürnberg,\\
Erlangen, Germany}
}

\specialpapernotice{(Invited Paper)}

\maketitle
\copyrightnotice

\begin{abstract}
In this paper, we propose a data augmentation framework for Optical Character Recognition (OCR). The proposed framework is able to synthesize new viewing angles and illumination scenarios, effectively enriching any available OCR dataset. Its modular structure allows to be modified to match individual user requirements. The framework enables to comfortably scale the enlargement factor of the available dataset. Furthermore, the proposed method is not restricted to single frame OCR but can also be applied to video OCR. We demonstrate the performance of our framework by augmenting a 15\% subset of the common Brno Mobile OCR dataset. Our proposed framework is capable of leveraging the performance of OCR applications especially for small datasets. Applying the proposed method, improvements of up to 2.79 percentage points in terms of Character Error Rate (CER), and up to 7.88 percentage points in terms of Word Error Rate (WER) are achieved on the subset. Especially the recognition of challenging text lines can be improved. The CER may be decreased by up to 14.92 percentage points and the WER by up to 18.19 percentage points for this class. 
Moreover, we are able to achieve smaller error rates when training on the 15\% subset augmented with the proposed method than on the original non-augmented full dataset.
\end{abstract}
%
%
%
%
\IEEEpeerreviewmaketitle
\section{Introduction}
Optical Character Recognition (OCR) gathers more and more attention with the ongoing digitalization of our daily life. Rapid advances in machine learning and especially neural networks spur the development of increasingly robust OCR applications \cite{Borisyuk2018}. The latest smartphones even enable to use efficient implementations of neural network applications \cite{Wang2020}, which further advances the wide use of inexpensive OCR applications. Such applications include digitizing printed, handwritten, and even historic documents \cite{Martinek2019}. Even though the available algorithms already achieve high accuracy, there is still room for further improvement. The performance of existing methods critically depends on the size of a labelled training dataset. However, the creation of large-scale datasets is a very elaborate and therefore costly work. In our paper, we present a framework that enables to synthesize new camera views, illumination scenarios, lens distortions and camera sensors. We also show that data from this framework boosts the performance of OCR applications. \par
The paper is structured as follows: In the next Section, the related state-of-the-art work is presented. In Section~\ref{sec:our_framework}, we introduce the proposed framework. Thereafter, we describe the experimental setup for evaluating the performance of the framework. The results of these tests are evaluated and discussed in Section~\ref{sec:evaluation}. We conclude the paper in Section~\ref{sec:conclusion}.
\vspace{-0.2cm}
\section{Related Work}
The used dataset is crucial for neural network training. A good performance of the final OCR system can only be achieved if the available training dataset represents the use case well. Also, distortions that occur in the test set should be included into the training dataset in order to train the network on handling these distortions. Another constraint is the size of the training dataset. The dataset has to be large enough in order to prevent overfitting \cite{Long2020}. A common method that can alleviate this problem to a certain degree is data augmentation \cite{Simard2003}. This technique is applied in nearly all neural network trainings and aims at increasing the diversity of the used dataset. Thereby, the overfitting problem can be alleviated and the trained network becomes more robust against distortions. Often used techniques for data augmentation are, e.g. rotation, translation, cropping, or zooming \cite{Joseph2021}. \par
As it is a quite common problem that available datasets have to be augmented in order to achieve satisfying results, more advanced methods evolved. An example for such a method is given in \cite{Storchan2019} where a Generative Adversarial Network (GAN) is used to augment a dataset for the training of an OCR system. The GAN adds artifacts that often occur within faxes to the original dataset. A drawback of this method is that the GAN itself has to be trained beforehand. Therefore, an unlabelled synthetic dataset has to be created \cite{Storchan2019}. While generating the synthetic dataset, the language model has to be considered. Moreover, an additional font classifier has to be trained. This classifier is trained in a supervised manner on a mixed real and synthetic dataset requiring a ground truth font labelling for the training dataset~\cite{Storchan2019}. In contrast, our novel framework proposed in the next section requires no training of auxiliary networks and no acquisition of additional datasets.
\begin{figure}[t]
	\centering
	\includegraphics[width=0.33\textwidth]{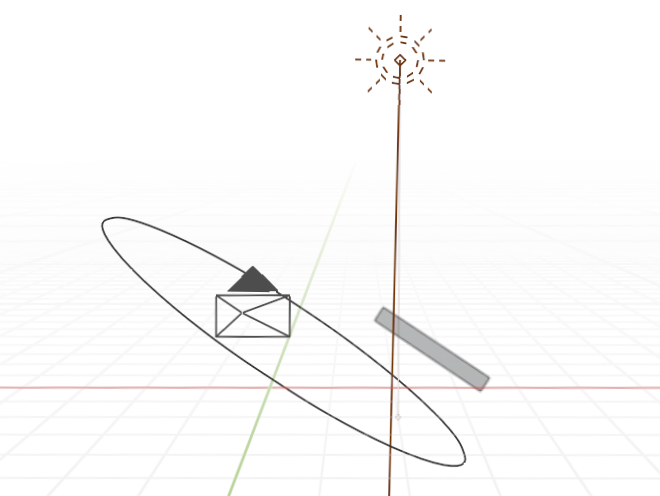}
	\caption{Example of a 3D scene generated with our framework for rendering augmented versions of the original sample. The scene contains a camera (black rectangle with filled black triangle on top) that acquires the augmented images and a light source (brown spot light) steering the illumination of the scene. Furthermore, a single text line from the original dataset is included (schematically shown as filled gray rectangle), as well as a rotated curve (black ellipse) on which the text line moves through the scene.}
	\label{fig:blender_scene}	
	\vspace{-0.4cm}
\end{figure}%
\section{Proposed Data-Augmentation Framework}
\label{sec:our_framework}%
Our proposed framework is intended as a preprocessing pipeline that enables the user to increase the number of available training samples by an almost arbitrary factor. The framework combines different illumination scenarios, different cameras, as well as different perspectives. \par 
The backbone of our framework is the open source 3D computer graphics software Blender (v2.83) \cite{BlenderFondation2020}. The whole augmentation process is steered by Python scripts configuring the individual Blender scenes. These scripts enable a full automation of the framework such that user inputs via the graphical user interface are not required. Thus, whole datasets can be augmented fully automatically in order to boost the performance of the final OCR application. The novel generated views of the training dataset are mainly specified by three aspects. These are the illumination of the scene, the specified camera, and a motion trajectory along which the training sample moves through the scene.\par 
\begin{figure}[t]
	\centering
	\resizebox{0.4\textwidth}{!}{
 	\begin{tikzpicture}

\node[align=center] at (5,7) (In) {Original Input Image\\of a Text Line};
\node[draw] at (5,6) (Blender) {3D Scene Generation};
\node[draw, align = center] at (5,5) (Render) {Rendering};
\node[draw] at (5,4) (Rot) {Rotation Compensation};
\node[draw] at (5,3) (Crop) {Extraction of Text Line};
\node[draw] at (5,2) (Resize) {Image Resizing};
\node[draw] at (5,0) (OCR) {Training of the OCR Application};

\node[align=center] at (9.5,7) (Light) {Camera, Light, and\\Trajectory Parameters};
\node[align=center] at (1,7) (Move) {Output Resolution,\\Scene Length};

\draw[dotted] (2.2,6.5) rectangle (7.8,1.3) node at (6.5,1.45) []{Data Augmentation};

\coordinate[below of=Light] (a1);  
\coordinate[below of=Move] (e1);

\draw [->] (In) -- (Blender);
\draw [->] (Blender) -- (Render); 
\draw [->] (Render) -- (Rot); 
\draw [->] (Rot) -- (Crop);
\draw [->] (Crop) -- (Resize);
\draw [->] (Resize) --  (OCR) node at (6.8,0.65) []{Set of Augmented Images};

\draw [->] (Light) |- (a1) -- (Blender);
\draw [->] (Move) |- (e1) -- (Blender);
\end{tikzpicture}}
	\caption{Flow graph demonstrating the workflow during the data augmentation process with the proposed framework. The proposed framework is denoted by the dashed box holding all steps performed during its application.}
	\label{fig:flow_graph}
	\vspace{-0.3cm}	
\end{figure}
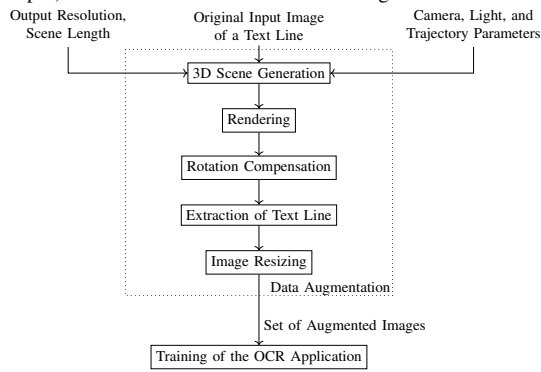 %
The augmentation process begins with an empty scene. Into this scene a camera is imported. The used cameras may vary in position, sensor size, lens, and its focal length. This option allows to simulate a large variety of different cameras and fields of view. Besides the camera, also the illumination of the scene may be varied, and chosen from a large variety of different light sources. The choice ranges from sunlight that uniformly illuminates the whole scene over area- and spotlights to a point light source. All these different light sources can be configured individually and may even be combined. That enables the user to remap several scenarios matching ones individual needs. Besides camera and illumination, the text that should be augmented has a central role in the scene. The original training sample from the available dataset is imported as an image plane into the scene. This image plane is then moved through the scene along an individually parametrized trajectory in front of the camera. An example of a complete scene generated with our framework can be seen in Figure~\ref{fig:blender_scene}. The movement along the curve results in a sequence of images holding different perspective views and illuminations of the original text sample. \par
By adjusting the sequence length, one can steer the factor by which the original dataset is increased. Before the augmented text lines can be used for training the OCR application, the black scene background has to be removed. Therefore, the bounding box coordinates of the text lines are saved during the rendering process. With these coordinates, the text line can be extracted from the rendered frame. In a first step, it is checked whether the text line is completely contained in the output frame. Afterwards, the rotated minimal enclosing rectangle of the bounding box is searched. Based on this rectangle, a rotation matrix is estimated. The image is then warped using the rotation matrix. From this warped image, the text line is extracted. During this process only the rotational transformation of the augmentation is inverted, all other transformational aspects such as shearing are still included in the augmented version of the text line. As a final processing step, the augmented images are resized using bicubic interpolation. The augmented images are scaled such that their resulting height is in line with the original dataset. This prevents drops in the recognition rate due to varying image sizes. During the rescaling process the aspect ratio of the augmented text lines remains unchanged. A summary of our proposed framework is given in Figure~\ref{fig:flow_graph} showing the single steps and the processing order as described above.%
\section{Experimental Setup}
\label{sec:test_setup}%
\begin{figure}[t]
	\centering
	\includegraphics[width=0.35\textwidth]{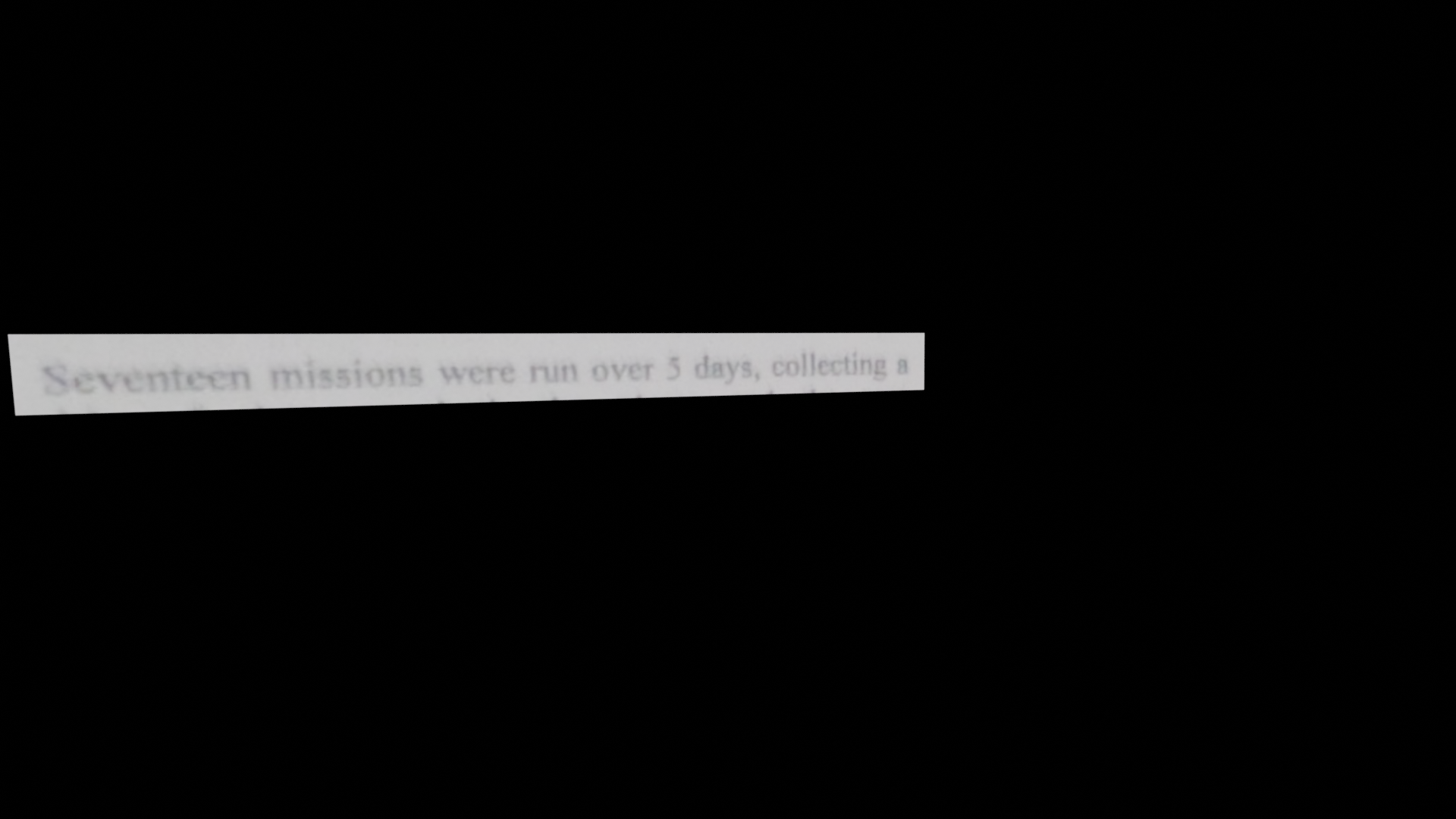}
	\caption{Example of an output frame showing a text line augmented using the proposed framework before rotation compensation and cropping.}
	\label{fig:aug_img}	
	\vspace{-0.4cm}
\end{figure}%
For the validation of our framework we used the Brno Mobile OCR Dataset \cite{Kiss2019}. This dataset holds excerpts from scientific papers that were acquired using different mobile devices, i.e. smartphones or tablets. The dataset provides single lines of English text that are fully labelled. The ground truth labels were obtained by a combination of Tesseract and ABBYY FineReader \cite{Kiss2019}. All text lines within the dataset are rectified. The dataset is split up into three subsets - easy, medium and hard - depending on the difficulty of the OCR task \cite{Kiss2019}. We used the single text lines from the Brno dataset as input images for our framework. In order to generate comparative results, we evaluated the performance of our OCR application trained on a subset holding 15\% of the original dataset compared to the performance of the application when it is trained on the same subset but enlarged by our proposed augmentation method. Moreover, we compare the results obtained using the proposed method to a network trained on the full original dataset.\par
Within the experiments conducted here, we used sun light as a homogeneous light source. We used four different cameras during the augmentation. The parameters were chosen such that a broad variety of common camera types is modelled e.g. mobile phones or full-frame consumer cameras. The parameters specifying each camera type are especially the sensor size and focal length of the lens. Furthermore, the position of each camera in the scene is chosen such that it matches the common application of the specific camera type. The single text lines are moved on a circular curve in front of the camera. The image of the text line is oriented such that it always faces the camera. The center of the curve is fixed for all scenes whereas the radius is varied randomly. Additionally, the curve is rotated around the axis perpendicular to the camera plane. The rotation angle is randomly chosen in the range from $-45$ to $45$ degree. During the augmentation process we generate ten frames for each scene using Blender's Cycles GPU render engine. Figure~\ref{fig:aug_img} shows an example frame of a text line rendered by our proposed framework. \par%
\begin{figure}[t]
	\centering
	\includegraphics[width=0.4\textwidth]{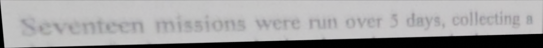}
	\caption{Example of a cropped text line after all augmentation methods were applied. The shearing from the augmentation is still visible after the rotation-compensation.}
	\label{fig:aug_line_cropped}
	\vspace{-0.2cm}	
\end{figure} %
Another parameter that might be varied within the framework is the output resolution of the renderer. For our experiments we chose a high render resolution of $1920 \times 1080$ pixels as our goal was to synthesize new additional viewing angles while affecting the quality of the final text lines as little as possible. By decreasing the output resolution, the processing time can be sped up drastically. But these time savings come at the price of lower image quality which might impair the performance of the trained network. Another factor heavily influencing the processing time is the sequence length, as this steers the number of frames rendered per sequence. While choosing a shorter sequence length speeds up the augmentation process, a longer sequence length generates more novel viewing angles and illuminations resulting in a bigger expansion of the original dataset. The parameter has to be chosen according to the specific use case, depending also on external factors as time limitations, available hardware, and especially the size of the original dataset.\par
\begin{figure}[t]
	\centering
	\includegraphics[width=0.4\textwidth]{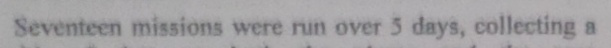}
	\caption{Example of the original text line before our augmentation framework was applied.}
	\label{fig:orig_line}
	\vspace{-0.5cm}	
\end{figure}

 
Before the renderings as depicted in Figure~\ref{fig:aug_img} can be used for training the OCR application, the line has to be extracted from the frame. Therefore, we use the correction and cropping procedure explained in Section~\ref{sec:our_framework}. During the correction we only cope with the rotational transformation in the renderings. Figure~\ref{fig:aug_line_cropped} shows a text line extracted from the frame depicted in Figure~\ref{fig:aug_img}. While the rotation was corrected, the shearing of the text line is still visible. Moreover, the final resolution of the augmented text line differs from the original one. While the line was resized such that it matches the original height, the augmented image of the line is narrower due to the geometric transformation applied during the augmentation process. The visual quality of the text line, does not experience substantial losses when compared to the original text line depicted in Figure~\ref{fig:orig_line}. Nevertheless, the contrast of the augmented text line decreases due to the additional illumination during the augmentation process.\par %
\section{Evaluation of the Proposed Framework}
\label{sec:evaluation}
For the evaluation of our data augmentation framework we employed the commonly used Tesseract framework \cite{Smith2007}. In its current version 4.1.1, Tesseract uses a LSTM network architecture, which is the current state of the art in OCR applications \cite{Tesseract2019}. For all evaluations, the network was trained independently from scratch. We trained the networks for two epochs each. The training dataset varies with each experiment as stated below. The test dataset is identical in all conducted experiments. The augmentation steps are only applied during the training of the networks. For testing, we use data in the original form as provided in the Brno Mobile OCR dataset \cite{Kiss2019}. This guarantees a fair comparison and prevents a privileged handling of networks favouring the transformations introduced by the data augmentation procedure. \par
As in \cite{Kiss2019}, we evaluate the performance of all our trained configurations in terms of Character Error Rate (CER) and Word Error Rate (WER). The CER is computed for every line as 
\begin{equation}
\label{eq:CER}
\mathrm{CER}=\frac{d_{\mathrm{C}}\left(\hat{L},L\right)}{N_{\mathrm{C}}},
\end{equation}
i.e. the Levenshtein distance on character level $d_{\mathrm{C}}$ between the recognized line $\hat{L}$ and the ground-truth line $L$ is normalized by the total number of characters in the ground-truth text line $N_{\mathrm{C}}$. Analogously, the WER is given as 
\vspace{-0.2cm}
\begin{equation}
\label{eq:WER}
\mathrm{WER}=\frac{d_{\mathrm{W}}\left(\hat{L},L\right)}{N_{\mathrm{W}}},
\end{equation}
where the Levenshtein distance on word level $d_{\mathrm{W}}$ is considered and normalized by the number of words in the ground truth line $N_{\mathrm{W}}$.\par%
In order to assess the effectiveness of our proposed augmentation framework, we trained the networks on 60,000 samples of the original training dataset from \cite{Kiss2019}. This is about $15\%$ of the original size of the dataset and shows the ability of our framework to increase the recognition rate even for small datasets. Our training dataset holds 2,400 text lines from the class 'hard', 14,100 text lines from the class 'medium' and 43,500 text lines from the class 'easy'. This distribution over the three classes, matches the distribution from the original Brno Mobile OCR training dataset \cite{Kiss2019}. \par 
The baseline of our evaluations is the network trained on the subset holding only the original samples. This baseline network is already able to achieve an overall CER of $5.09\%$ and an overall WER of $15.60\%$. To further decrease the error rates, we train the network on the subset enlarged with our proposed framework. For these experiments, we investigated enlargement factors from one to seven. An enlargement factor of one means that only original training samples are used and therefore serves as reference. An enlargement factor of two means, that we add one augmented version of every original sample to the dataset, i.e. the size of the dataset is increased by a factor of two. The datasets with the remaining enlargement factors are generated analogously. \par
Figure~\ref{fig:CER_plot} shows the CER over the enlargement factor of the dataset. We observe, that the CER decreases further with a growing number of augmented samples included into our training dataset. The same holds for the WER that is plotted over the enlargement factor of the dataset in Figure~\ref{fig:WER_plot}. The shape of the WER curves is similar to the CER, as the two metrics are related. Nevertheless, it can be observed that the WER is always higher than the CER. The reason for this is that the WER is scaled to the number of words per line, whereas the CER is scaled to the number of characters per line. Hence, the single errors affect the WER more strongly than the CER. The exact CER obtained on the test dataset are given in Table~\ref{tab:res_diff_CER}. The overall error rates are calculated as an average of the three classes weighted by the number of samples per class. The table shows the results for all examined enlargement factors. The best results are given in bold. In the bottom line the error rates for the network trained on the original full dataset holding 403,744 samples are given. All error rates are given with respect to the results obtained with the baseline version trained on the non-augmented subset of the original dataset only. Analogously, Table~\ref{tab:res_diff_WER} shows the WER for the conducted experiments.\par
Analyzing the numbers given in Table~\ref{tab:res_diff_CER} and \ref{tab:res_diff_WER} it can be observed that the error rates drop especially for the classes 'medium' and 'hard'. By using the proposed data augmentation framework we add more challenging perspectives to the training datasets. This increases the performance of the trained OCR network on challenging text lines in the final test set. We are able to show improvements of up to 14.92 percentage points in terms of CER, and up to 18.19  percentage points in terms of WER for the class 'hard'. We are even able to outperform the network trained on the full dataset. Training the network on a 15\% subset and using our proposed data augmentation method with an enlargement factor of at least three holds better results for all classes in terms of WER and CER. 
\begin{table}[ht]
\centering
\caption{Character error rate for the augmented training datasets and full dataset w.r.t. original 15\% subset.}
\label{tab:res_diff_CER}
\begin{tabular}{|l|c|c|c|c|} \hline
& Easy & Medium & Hard & Overall \\ \hline
Reference & $1.35\%$  & $11.81\%$ & $40.91\%$  & $5.09\%$  \\ \hline
2 & $-0.73\%$ & $-4.50\%$ & $-8.83\%$  & $-1.87\%$ \\ \hline
3 & $-0.80\%$ & $-5.21\%$ & $-10.16\%$ & $-2.13\%$ \\ \hline
4 & $-0.91\%$ & $-5.75\%$ & $-10.13\%$ & $-2.33\%$ \\ \hline
5 & $-0.94\%$ & $-5.84\%$ & $-10.96\%$ & $-2.40\%$ \\ \hline
6 & $-0.98\%$ & $-6.72\%$ & $\mathbf{-14.92\%}$ & $-2.77\%$ \\ \hline
7 & $\mathbf{-0.99\%}$ & $\mathbf{-6.84\%}$ & $-14.64\%$ & $\mathbf{-2.79\%}$ \\ \hline \hline
Full Dataset & $-0.73\%$  & $-4.53\%$  & $-10.40\%$  & $-1.91\%$ \\ \hline
\end{tabular}
\end{table}%
\begin{table}[ht]
\centering
\caption{Word error rate for the augmented training datasets and full dataset w.r.t. original 15\% subset.}
\label{tab:res_diff_WER}
\begin{tabular}{|l|c|c|c|c|} \hline
& Easy & Medium & Hard & Overall \\ \hline
Reference & $6.28\%$  & $36.10\%$  & $79.90\%$  & $15.60\%$\\ \hline
2 & $-3.11\%$ & $-11.16\%$ & $-9.68\%$  & $-5.17\%$\\ \hline
3 & $-3.49\%$ & $-12.74\%$ & $-10.74\%$ & $-5.85\%$\\ \hline
4 & $-3.95\%$ & $-14.25\%$ & $-10.07\%$ & $-6.51\%$\\ \hline
5 & $-4.10\%$ & $-14.75\%$ & $-12.08\%$ & $-6.80\%$\\ \hline
6 & $-4.36\%$ & $-17.52\%$ & $\mathbf{-18.19\%}$ & $-7.84\%$\\ \hline
7 & $\mathbf{-4.38\%}$ & $\mathbf{-17.77\%}$ & $-17.44\%$ &  $\mathbf{-7.88\%}$\\ \hline \hline
Full Dataset & $-2.99\%$  & $-10.68\%$  & $-10.49\%$  & $-5.00\%$ \\ \hline
\end{tabular}
\vspace{-0.2cm}
\end{table}%
\begin{figure}[t]
	\centering
	\resizebox{0.4\textwidth}{!}{
%
%
\definecolor{mycolor1}{rgb}{0.00000,0.44700,0.74100}%
\definecolor{mycolor2}{rgb}{0.85000,0.32500,0.09800}%
\definecolor{mycolor3}{rgb}{0.92900,0.69400,0.12500}%
\definecolor{mycolor4}{rgb}{0.49400,0.18400,0.55600}%
\begin{tikzpicture}

\begin{axis}[%
width=4.521in,
height=3in,
at={(0.758in,0.481in)},
scale only axis,
xmin=1,
xmax=7,
xtick distance = 1, 
xlabel style={font=\color{white!15!black}},
xlabel={\LARGE{Dataset enlargement factor}},
ymin=0,
ymax=50,
ytick distance = 5, 
ylabel style={font=\color{white!15!black}},
ylabel={\LARGE{CER in \%}},
axis background/.style={fill=white},
legend style={legend cell align=left, align=left, draw=white!15!black},
mark size=1mm
]
\addplot [color=mycolor1, mark=*, line width = 0.7mm]
  table[row sep=crcr]{%
1	5.087865099485287\\
2	3.220511830682953\\
3	2.961374548667127\\
4	2.757908120150573\\
5	2.686923254206038\\
6	2.321441960513175\\
7	2.296203426288699\\
};
\addlegendentry{Overall}


\addplot [color=mycolor2, mark=*, dotted, line width = 0.7mm]
  table[row sep=crcr]{%
1	1.35\\
2	0.62\\
3	0.55\\
4	0.44\\
5	0.41\\
6	0.37\\
7	0.36\\
};
\addlegendentry{Easy}

\addplot [color=mycolor3, mark=*, dash pattern=on 1pt off 3pt on 3pt off 3pt, line width = 0.7mm]
  table[row sep=crcr]{%
1	11.81\\
2	7.31\\
3	6.60\\
4	6.06\\
5	5.97\\
6	5.09\\
7	4.97\\
};
\addlegendentry{Medium}

\addplot [color=mycolor4, mark=*, dash pattern=on 3pt off 6pt on 6pt off 6pt, line width = 0.7mm]
  table[row sep=crcr]{%
1	40.91\\
2	32.08\\
3	30.75\\
4	30.78\\
5	29.95\\
6	25.99\\
7	26.27\\
};
\addlegendentry{Hard}

\end{axis}
\end{tikzpicture}%
}
	\caption{CER evaluated on the test set for different enlargement factors of the training dataset. An enlargement factor of one corresponds to only original samples used and therefore serves as reference. Besides the overall CER, the CER is given for the sub-classes 'easy', 'medium' and 'hard' as defined in \cite{Kiss2019}.}
	\label{fig:CER_plot}
	\vspace{-0.1cm}	
\end{figure}
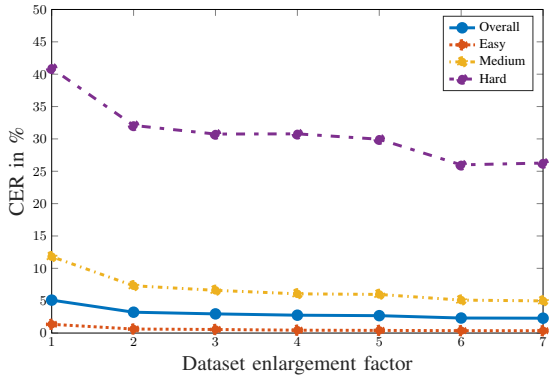%
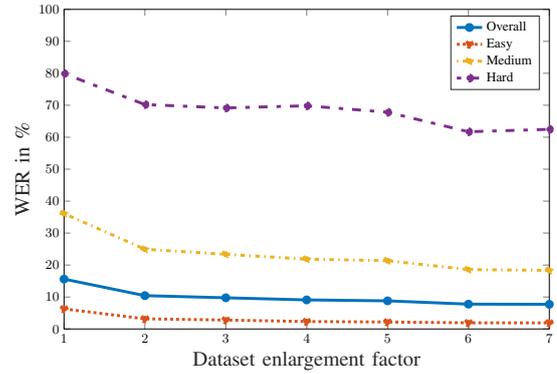
\begin{figure}[t]
	\centering
	\resizebox{0.4\textwidth}{!}{
%
%
\definecolor{mycolor1}{rgb}{0.00000,0.44700,0.74100}%
\definecolor{mycolor2}{rgb}{0.85000,0.32500,0.09800}%
\definecolor{mycolor3}{rgb}{0.92900,0.69400,0.12500}%
\definecolor{mycolor4}{rgb}{0.49400,0.18400,0.55600}%
\begin{tikzpicture}

\begin{axis}[%
width=4.521in,
height=3in,
at={(0.758in,0.481in)},
scale only axis,
xmin=1,
xmax=7,
xtick distance = 1, 
xlabel style={font=\color{white!15!black}},
xlabel={\LARGE{Dataset enlargement factor}},
ymin=0,
ymax=100,
ytick distance = 10, 
ylabel style={font=\color{white!15!black}},
ylabel={\LARGE{WER in \%}},
axis background/.style={fill=white},
legend style={legend cell align=left, align=left, draw=white!15!black}
]
\addplot [color=mycolor1, mark=*, line width = 0.7mm]
  table[row sep=crcr]{%
1	15.602005070292696\\
2	10.428544211415842\\
3	9.751182300069141\\
4	9.089751862948452\\
5	8.796361488822310\\
6	7.763625067219788\\
7	7.717341361296765\\
};
\addlegendentry{Overall}

\addplot [color=mycolor2, mark=*, dotted, line width = 0.7mm]
  table[row sep=crcr]{%
1	6.28\\
2	3.17\\
3	2.79\\
4	2.33\\
5	2.18\\
6	1.92\\
7	1.90\\
};
\addlegendentry{Easy}

\addplot [color=mycolor3, mark=*, dash pattern=on 1pt off 3pt on 3pt off 3pt, line width = 0.7mm]
  table[row sep=crcr]{%
1	36.10\\
2	24.94\\
3	23.36\\
4	21.85\\
5	21.35\\
6	18.58\\
7	18.33\\
};
\addlegendentry{Medium}

\addplot [color=mycolor4, mark=*, dash pattern=on 3pt off 6pt on 6pt off 6pt, line width = 0.7mm]
  table[row sep=crcr]{%
1	79.90\\
2	70.22\\
3	69.16\\
4	69.83\\
5	67.82\\
6	61.71\\
7	62.46\\
};
\addlegendentry{Hard}

\end{axis}
\end{tikzpicture}%
}
	\caption{WER evaluated on the test set for different enlargement factors of the training dataset. An enlargement factor of one corresponds to only original samples used and therefore serves as reference. Besides the overall WER the WER is additionally given for the sub-classes 'easy', 'medium' and 'hard'.}
	\label{fig:WER_plot}
	\vspace{-0.3cm}
\end{figure} %
%
%
\section{Conclusion}
\label{sec:conclusion}
In this paper, we introduced a novel framework for data augmentation in the context of Optical Character Recognition applications. Our proposed framework is based on the open source 3D computer graphics render software Blender. It enables the user to enrich an existing dataset by a manifold of additional viewing angles, illumination scenarios, camera types and lenses. Due to its modular structure it can be modified to match individual user requirements. The framework allows to comfortably scale the factor by which the available dataset shall be enlarged. Furthermore, by scaling the render resolution the framework can be fitted easily to match the individual quality requirements and adapt to the personal hardware restrictions. \par
Using the well known Tesseract v4.1.1 LSTM implementation as an exemplary OCR application, our proposed framework is able to demonstrate that the recognition rate can be improved significantly. The overall CER can be decreased by 2.79 percentage points when the original dataset is enlarged by a factor of seven. The overall WER can be decreased by up to 7.88 percentage points. The usage of the proposed data augmentation framework is especially beneficial for challenging application cases. %
Moreover, it was shown that augmenting a 15\% subset with the proposed framework yields better results than training the network on the non-augmented full dataset.
\section*{Acknowledgment}
We gratefully acknowledge support by the German Federal Ministry of Educationand Research (BMBF) under Grant No. 13N15319.
\bibliographystyle{IEEEtran}
\bibliography{ocr_literature}
\end{document}